\documentclass[11pt]{article}

\usepackage[final]{acl}

\usepackage{times}
\usepackage{latexsym}
\usepackage{comment}
\usepackage[T1]{fontenc}

\usepackage[utf8]{inputenc}

\usepackage{microtype}

\usepackage{inconsolata}

\usepackage{graphicx}

\usepackage{booktabs}
\usepackage{fancyvrb}
\usepackage{fvextra}

\title{FLANS at SemEval-2026 Task 7: \\ RAG with Open-Sourced Smaller LLMs for Everyday Knowledge\\ Across Diverse Languages and Cultures}

\author{
           Liliia Bogdanova$^{1,\dagger}$, Shiran Sun$^{2,\dagger}$, \textbf{Lifeng Han}$^{*3,4}$ \\ \textbf{Natalia Amat Lefort}$^{3}$, \textbf{Flor Miriam Plaza-del-Arco}$^{3}$ 
              \vspace*{0.075cm}
\\            $^1$ Insilico Medicine AI Limited
            $^2$ University of Groningen
  \\          $^3$  LIACS, Leiden University
            $^4$  Leiden University Medical Center  
      \\   $^\dagger$LB and SS: \textit{co-first} alphabet order $^*$LH: \textit{Corresponding: l.han@liacs.leidenuniv.nl} 
}

\begin{document}
\maketitle
\begin{abstract}
This system paper describes our participation in the SemEval-2025 Task-7 ``Everyday Knowledge Across Diverse Languages and Cultures''. We attended two subtasks, i.e., Track 1: Short Answer Questions (SAQ), and Track 2: Multiple-Choice Questions (MCQ).
The methods we used are retrieval augmented generation (RAGs) with open-sourced smaller LLMs (OS-sLLMs). 
To better adapt to this shared task, we created our own culturally aware knowledge base (CulKBs) by extracting Wikipedia content using keyword lists we prepared. We extracted both culturally-aware wiki-text and country-specific wiki-summary. 
In addition to the local CulKBs, we also have one system integrating live online search output via DuckDuckGo.
Towards better privacy and sustainability, we aimed to deploy smaller LLMs (sLLMs) that are open-sourced on the Ollama platform.
We share the prompts we developed using refinement techniques and report the learning curve of such prompts.
The tested languages are English, Spanish, and Chinese for both tracks.
Our resources and codes are shared via \url{https://github.com/aaronlifenghan/FLANS-2026}
\end{abstract}

\section{Introduction}
We present the system report for SemEval Shared Task 7. Our contributions are:
(1) We propose a sustainable and locally deployable RAG framework using open-sourced small language models for multilingual cultural question answering.
(2) We construct a multilingual cultural knowledge base from Wikipedia and curated facts to support culturally grounded QA.
(3) We perform a prompt ablation study demonstrating the importance of structured prompt design for multilingual factual QA.
(4) We introduce a cascaded retrieval and model routing strategy that improves answer grounding while maintaining efficiency.

\section{Related Work}



\textbf{I) RAG for Culture-Aware Generation}
LLMs have been known for their struggles when dealing with 
culturally specific content due to knowledge sparsity. To address this issue, there are related work using RAGs for {culturally}-aware text generation. 
For example, from \textbf{country} and \textbf{region} perspective,
\newcite{lee2025evaluating}  proposes a structured evaluation framework to assess how well LLMs represent and generate minority cultural knowledge — specifically focusing on {Taiwanese} Hakka culture. 
Other language and culture specific RAG systems including Islamic and Arabic
\cite{alan2025improving-isslamic,abdelaziz2025arabic,faruk2025adab}, Yoruba Culture and Language (African) \cite{joshua2024improving-african}, etc.
Similarly, focusing on specific domains such as \textbf{education}, LLMs' hallucination can perpetuate bias or outdated knowledge that are especially risky when generating learning content for vulnerable populations. \newcite{joseph2024retrieval-refugee} focused on  refugee learners and explored RAGs as a strategy to improve the quality, contextual relevance, and cultural sensitivity of educational content.

In a \textbf{multilingual} setting, while RAG helps in knowledge-intensive tasks, it can also propagate and amplify biases present in retrieved documents, especially in cross-lingual settings where source document quality varies.
\newcite{li-etal-2025-multilingual} contributed a new dataset BordIRLines, containing territorial dispute descriptions paired with relevant retrieved {\textit{Wikipedia}} documents (we also used for our own work in this shared task). It covers 49 languages, spanning many language families and resource levels, designed to test how well RAG systems handle culturally sensitive retrieval and generation across languages.
To address \textbf{cost} of human annotated benchmark, 
\newcite{zhang-etal-2025-culturesynth}  provided a scalable synthesis framework that combines hierarchical cultural knowledge with retrieval mechanisms for higher-quality, culturally grounded QA generation.
Unlike benchmarks that rely heavily on manual creation, CultureSynth demonstrates how taxonomy-guided RAG can produce large amounts of useful evaluation data across languages.
Evaluation of 14 popular LLMs using CultureSynth shows larger models perform better on cultural QA tasks.
A parameter threshold around ~3B is necessary for basic cultural competence.
Closely following this route, we aim to explore lower-cost, \textit{sustainable}, \textit{secure}, and smaller-sized LLMs on this task, for which we list some related literature below.

\textbf{II) Sustainable and Secure AI}
Researchers have questioned the assumption that larger LLMs are always superior.
Scaling-law studies by \newcite{kaplan2020scaling} showed diminishing performance returns as the model size increases, while, instead the bigger the better, \newcite{han2024neuralMT}  demonstrated that smaller, better-trained models can match and even exceed over-sized counterparts, on the domain-specific machine translation (MT) task. 
NLP Researchers have started to highlight the substantial carbon and energy costs of large-score LM training and advocate efficiency-aware solutions \cite{strubell2019energy}, e.g. model distillation, quantization, and lightweight open-source models \cite{jiao-etal-2020-tinybert,hu2023lora}.
Motivated by these findings, we explore open-sourced small language models (OS-sLLMs) and RAG to reduce computational cost, enable local deployment, and improve privacy 
on culturally grounded QA.






\begin{figure}[t]
    \centering
    \includegraphics[width=0.5\textwidth]{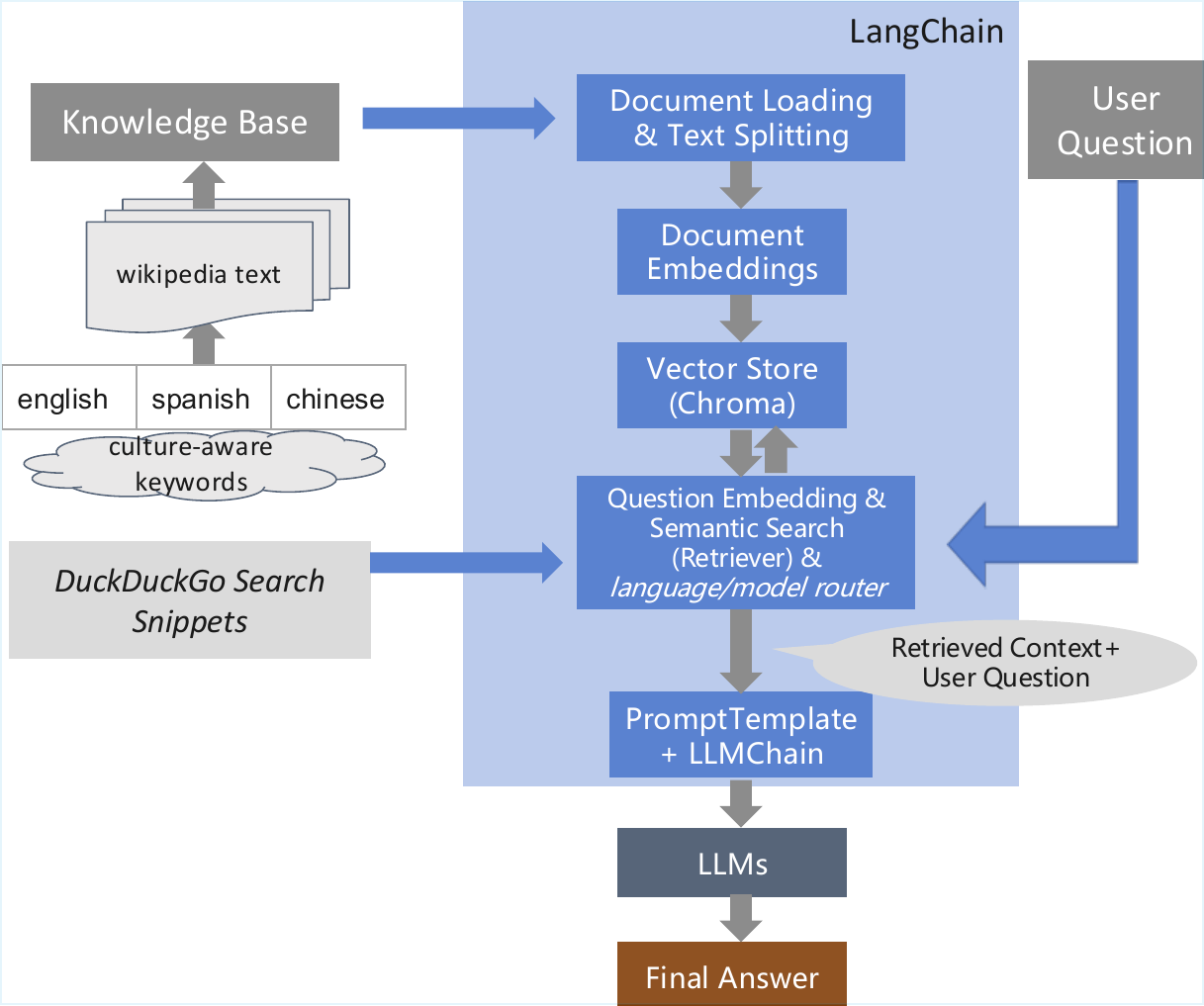}
    \caption{RAG pipeline. {\textit{italic}} indicating variations}
    \label{fig:rag_pipeline_integrated}
\end{figure}

\section{Methodology}
The methodology design of our system is shown in Figure \ref{fig:rag_pipeline_integrated}.
The system is based on a multilingual knowledge base (KB) in which documents are split into small text chunks, encoded using a multilingual embedding model, and indexed in a local Chroma vector store, as shown in Figure \ref{fig:rag_pipeline_integrated} (upper-half).
At inference time, the user question is embedded in the same vector space and matched against the index to retrieve the top-k most relevant passages via semantic search. The retrieved context is then combined with the original question and inserted into a structured prompt template, which is passed through the LLMChain to generate the final answer. The SLMs are instructed to output only a short factual response, ensuring concise and consistent outputs. short factual answer. This design enables the model to access culture-specific information while keeping the overall system light-weighted and compatible with local hardware.


The design of our system is \textbf{motivated} by the specific characteristics of the BLEnD shared task, which evaluates multilingual and culture-specific factual knowledge across diverse languages. Here we adopt a small-language-model (SLM) type. Small models alone lack sufficient parametric knowledge for culturally grounded questions, but their behaviors are expected to become more stable and predictable when combined with external evidence. 
Therefore, our system integrates:
(1) lightweight multilingual SLMs suitable for local deployment, (2) structured prompt design to enforce short, same-language answers, and (3) a (RAG) module using a curated external knowledge base. This architecture aims to balance accuracy, multilingual robustness, and computational feasibility while providing a transparent and extensible/adaptable framework for evaluating cultural question answering.
As a variation of our system (\textsc{RAG-web}), we added online search component via DuckDuckGo and language/model router function in \textit{italic}.

\subsection{Language and Model Selection}
We first define the language scope of our system, which directly informs our model selection under local deployment constraints.
Our development process focused on the three languages most relevant to our team and to the early stages of the BLEnD dataset: Chinese, English, Spanish. These languages were used to test model behavior, prompt stability, and retrieval quality in a multilingual setting. 

To identify a suitable model for the BLEnD short-answer questions, especially for Chinese queries, we first compared four small language models available in Ollama that match our local hardware constraints: Phi-4-Mini, Llama 3.2 (3B), DeepSeek-R1 (7B), and Gemma 3 (4B). We asked each model to generate answers for a subset of BLEnD SAQ items and evaluated them based on accuracy, response stability, and runtime efficiency. Among these candidates, Gemma 3 showed the most balanced performance, providing higher accuracy on Chinese factual questions and noticeably faster, more consistent output. Based on these observations, we selected Gemma 3 as the primary model for our \textsc{RAG-base} system.

In the \textit{variation} version (\textsc{RAG-web}), we explored two models \texttt{mistral:7b} and \texttt{deepseek-llm:67b}. The models are chosen dynamically using the language identifier embedded in each of the BLEnD questions’ ID. For the majority of inputs that are not Chinese, the \texttt{mistral:7b} model is used. This is because it strikes a good balance between multilinguality, inference speed, and local deployability. However, experiments have revealed that it is not very effective for Chinese queries in region-specific contexts.
For this problem, a \textit{deterministic routing} approach is adopted. If the language ID is one of the Chinese variants (such as \texttt{zh-CN}, \texttt{zh-TW}, \texttt{zh-SG}), the system switches to using the \texttt{deepseek-llm:67b} model. This is because it is much larger but performs better for Chinese factual reasoning and named entity recognition.

\subsection{Prompt Development}


We evaluate three prompt variants of increasing structural and cognitive constraint, denoted as Minimal Prompt (\textbf{MP}) and Refined Prompts (\textbf{RP-v1} and \textbf{RP-v2}).
MP served as a baseline and used a minimal instruction, asking the model to produce one short, correct answer in the same language as the question, without additional guidance on format or reasoning.
For Refined Prompts (RPs),  we explore different kinds of techniques such as  persona, format instruction, Chain-of-Thought (CoT) and perspective-aware prompting \cite{ren2025paisp,romero2025manchester}. Main design elements of RP-v1 include:
\begin{itemize}
    \item Assigning the model the persona of being a factual multilingual assistant for a question-answering benchmark.
    \item Enforcing a strict output format instruction that produces only one concise answer in the same language as the question, without any explanations or extra text.

\end{itemize}
The goal of RP-v1 is to make the model’s output more concise and consistent, and to reduce common errors such as unnecessary words or language switching, which small language models sometimes produce in multilingual settings.
RP-v2 introduces a perspective-aware step inspired by PA-ISP, incorporating implicit CoTs reasoning to guide the model’s internal analysis before producing the final answer~\cite{ren2025paisp}.
In this version, main design elements include, in addition to the persona and strict output format from RP-v1:
    a) Introducing a structured self-guided reflection phase before answering,
    b) Guiding the model to think internally in several structured steps, including question analysis, information focus, answer strategy, and error avoidance.

Among the Refined Prompts (RP-v1 and RP-v2), the difference is that RP-v2 adds an explicit internal analysis stage before answer generation. It encourages the model to examine the question from multiple perspectives, such as language identification and information type, before deciding on the final answer. 
This design aims to improve answer accuracy and consistency in multilingual settings, especially for short-answer questions where the model must select a single precise fact.

\subsection{Knowledge Base Constructions}

For \textbf{\textsc{RAG-base}} Knowledge Base Construction (\textbf{KBC}), we build a multilingual knowledge base in Chinese, English, and Spanish to support the question-answering task. The KB combines two sources: (1) Wikipedia content and (2) manually curated cultural facts. Wikipedia articles are selected to cover common types of general-knowledge questions, including national symbols, geography, history, culture, and society. For each topic, we extract sentences from introductory summaries, which typically contain clear and concise factual information. These segments, either sentences or short paragraphs, are used as individual KB entries, resulting in approximately 700 entries from Wikipedia. To ensure reliable coverage of important and frequently asked cultural facts, we additionally include around 200 manually curated statements, such as official currencies or well-known national associations. All KB entries are embedded using the OllamaEmbeddings model and indexed with Chroma for vector-based retrieval. \footnote{we share our KB of wiki extraction and manually written facts at our page \url{https://github.com/aaronlifenghan/FLANS-2026} }
For \textsc{RAG-base} system, if the model retrieve some knowledge, it will print the sentence out, otherwise it prints ``NULL".



\subsection{Creating Pseudo Ground Truth}
Since the official BLEnD gold answers cannot be used for training or tuning, we created a small pseudo ground truth set to support model development and system debugging. This set was constructed using two complementary methods. First, a small subset of Chinese, English and Spanish questions was manually annotated by native speakers or near-native speakers to ensure correctness. Second, we used GPT-4 to generate initial candidate answers, which were then manually checked and corrected via post-editing. This hybrid approach allowed us to build a reliable reference set without introducing data leakage from the official test labels. The pseudo ground truth was used only for internal evaluation of prompt quality, model selection, and RAG behavior, providing a controlled way to assess system changes during development.

\subsection{Language Routing}
Given BLEnD's wide range of languages and cultural variants, we use a deterministic language routing approach. The language code embedded in each question ID (e.g., \texttt{es-MX}, \texttt{en-SG}, \texttt{zh-CN}) determines both the country-level knowledge base and the LLM used for answer generation. This means that Spanish queries about Mexico pull Mexican cultural information, while English queries about Singapore pull Singapore-specific sources. For Chinese variants, this routing also activates the \texttt{DeepSeek} model, which tends to give more reliable results than the lighter alternatives.

\begin{table*}[t]
\centering
\small
\begin{tabular}{lcccccc}
\toprule
\textbf{Language} 
& \multicolumn{3}{c}{\textbf{Track 1 (SAQ)}} 
& \multicolumn{3}{c}{\textbf{Track 2 (MCQ)}} \\
\cmidrule(lr){2-4} \cmidrule(lr){5-7}
& \textbf{MP} & \textbf{RP-v1} & \textbf{RP-v2}
& \textbf{MP} & \textbf{RP-v1} & \textbf{RP-v2} \\
\midrule

English (en) 
& 17.14 & 24.29 & 37.14 
& 82.86 & 82.86 & 82.86 \\

Spanish (es) 
& 4.17 & 35.00 & 47.50
& 80.83 & 85.00 & 70.00 \\

Chinese (zh) 
& 27.14 & 41.43 & 48.57 
& 65.71 & 82.86 & 68.57 \\

\midrule
\textbf{Average} 
& 16.15 & 33.57 & 44.40
& 76.46 & 83.57 & 73.81 \\

\bottomrule
\end{tabular}
\caption{\textsc{RAG-base} Prompt ablation results by language and task.}
\label{tab:prompt_ablation_RAGbase}
\end{table*}

\begin{table*}[t]
\centering
\small
\begin{tabular}{lcccccc}
\toprule
\textbf{Language}
& \multicolumn{3}{c}{\textbf{Track 1 (SAQ)}} 
& \multicolumn{2}{c}{\textbf{Track 2 (MCQ)}} \\
\cmidrule(lr){2-4} \cmidrule(lr){5-6}
& \textbf{RP-v1}
& \textbf{RP-v2}
& \textbf{RP-v1 (no local DB)}
& \textbf{RP-v1}
& \textbf{RP-v1 (no local DB)} \\
\midrule
English (en)
& 16.67 & 0.00 & 16.67
& 83.33 & 83.33 \\

Spanish (es)
& 33.33 & 0.00 & 33.33
& 61.11 & 72.22 \\

Chinese (zh)
& 33.33 & 8.33 & 66.67
& 91.67 & 83.33 \\
\midrule
\textbf{Average}
& 27.78 & 2.78 & 38.89
& 78.70 & 79.63 \\
\bottomrule
\end{tabular}
\caption{\textsc{RAG-web} Prompt ablation results by langauge and task.}
\label{tab:prompt_ablation_expanded_liliia}
\end{table*}

\section{System Development and Validation}
\subsection{Experimental and Evaluation Setup}
For \textsc{RAG-base}:
we use Gemma-4B and run all experiments locally on a CPU-based laptop (Intel i5-10210U, 16 GB RAM).
No GPU or external computing infrastructure is involved.
This highlights the feasibility of lightweight and sustainable deployment.

For \textsc{RAG-web}:
we run experiments for MCQ questions locally using Ollama with Mistral and DeepSeek models on a MacBook Pro (Apple M1, 16 GB RAM, 512 GB SSD).
All computations were performed on CPU-based Apple Silicon without discrete GPU or external infrastructure.
This demonstrates that competitive multilingual cultural QA performance can be achieved using modest consumer hardware, supporting sustainable and accessible AI deployment.

We use the standard BLEnD \textbf{evaluation protocol} for system evaluation. For the Short Answer Question (SAQ) track, a system prediction is said to be correct if it matches any of the human reference answers for a given question. For the Multiple Choice Question (MCQ) track, correctness is measured by matching a system prediction with a ground truth label.
The final score is reported as accuracy averaged across all language tracks. More details about the evaluation protocol can be found in \cite{myung2024blend}. 

\subsection{Ablation Study of Prompts}

    

\begin{figure}[htbp]
    \centering
    \includegraphics[width=\columnwidth]{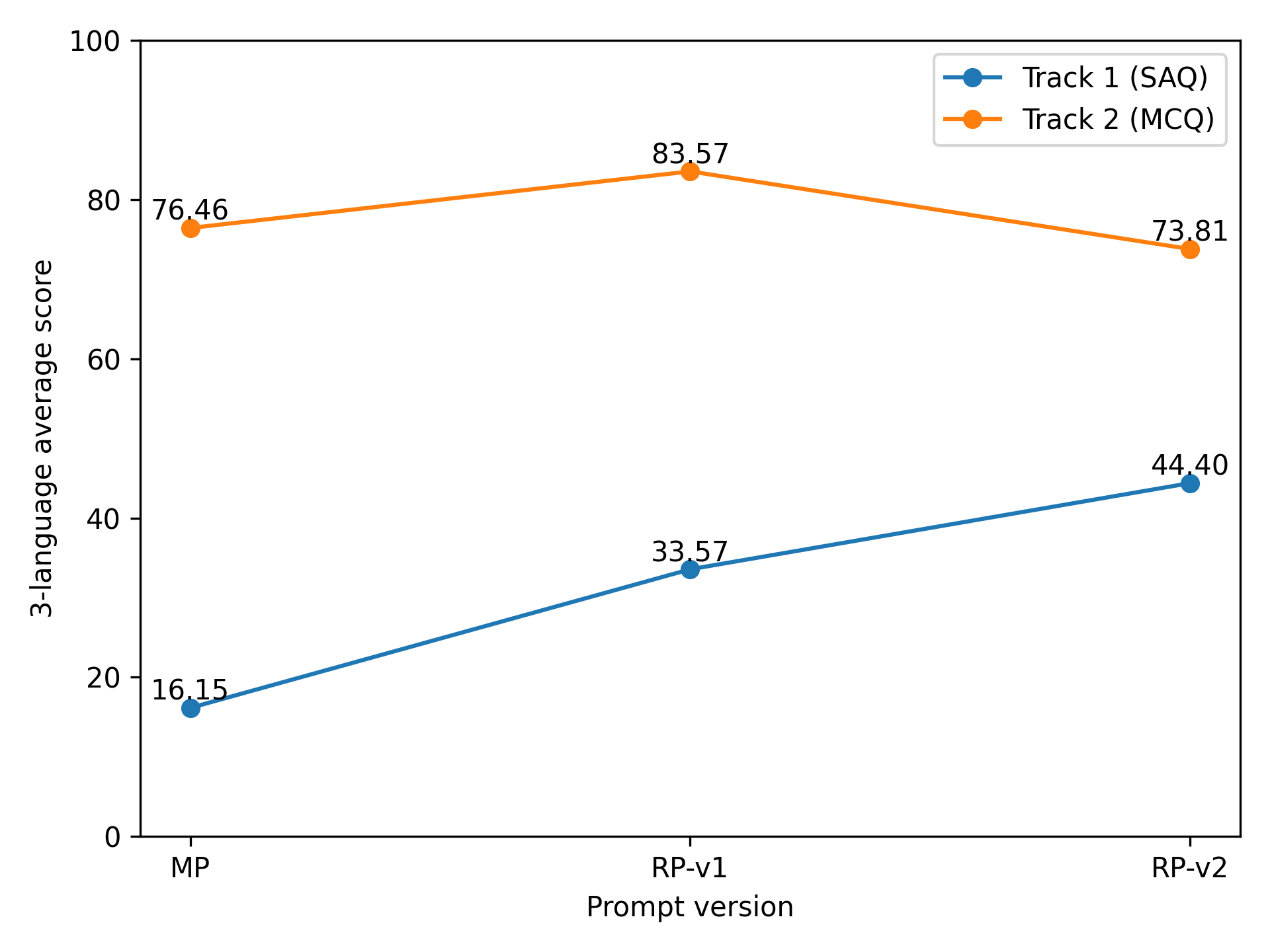}
    \caption{\textsc{RAG-base} Learning curves of three prompt ablation averaged over three-languages (en, es, zh).
    }
    \label{fig:prompt_learning_curve}
\end{figure}

Since our system only covers three languages (English, Spanish, and Chinese), the remaining languages contribute zero scores using the official evaluation platform, resulting in a lower overall score than the actual performance on the evaluated languages.
To obtain a clearer view, we report results for the selected languages only.
For each language, scores are averaged over regional variants (e.g., \texttt{zh-CN} and \texttt{zh-SG} for Chinese).
Detailed per-language results for both Track~1 (SAQ) and Track~2 (MCQ) under different prompt versions are shown in Table~\ref{tab:prompt_ablation_RAGbase} using our \textsc{RAG-base} system.
Figure~\ref{fig:prompt_learning_curve} further visualizes the three-language average scores across prompt versions.

For Track~1, performance consistently improved from MP to RP-v2, suggesting that both explicit formatting instruction and perspective-aware prompting are beneficial for open-ended short-answer generation, where the answer space is relatively unconstrained.
In contrast, Track~2 performance peaked at RP-v1 and slightly declined with RP-v2. This suggests that for multiple-choice questions, where the answer space is already tightly constrained by predefined options, additional internal reflection may introduce unnecessary cognitive overhead and interfere with option-level matching.

Overall, the prompt ablation suggests that most performance gains come from explicit external constraints on output format and multilingual behavior.
While self-guided reflection can further improve performance on open-ended short-answer questions, its benefits are limited for tasks with a tightly constrained answer space, such as multiple-choice questions. Based on these observations, we adopt RP-v1 as the default prompt in our \textsc{RAG-base} system, as it provides a strong balance between performance, simplicity, and robustness across both tracks.

For \textbf{\textsc{RAG-web}}, as shown in Table \ref{tab:prompt_ablation_expanded_liliia},
scores are weighted averages over regional variants based on the number of evaluation questions.
The evaluation on Chinese has much improved scores for Track 1 (SAQ) using RP-v1 achieving 66.67 vs the highest score 48.57 from \textsc{RAG-base}, although the other two languages have dropped scores.
Here, ``no local DB'' means only using online search duckduckgo and wikipedia. 
For Track2, using local database, Chinese further improved the score from 83.33 to 91.67, but Spanish score decreased 10 absolute points (72.22 to 61.11), which means that the Spanish DB might have introduced noise. This leads to future work to explore DB quality control for RAG systems.
Overall, we observe that performance on MCQ is consistently higher than on SAQ across all languages, reflecting the relative difficulty of free-form short-answer generation.
Chinese achieves the strongest MCQ performance, benefiting from language-specific model routing to a larger LLM, while Spanish shows comparatively stronger gains on SAQ after aggregation.




\subsection{Submission to SemEval-Test}

For official submission to SemEval Task 7, we submitted three systems using \textsc{RAG-base}, \textsc{RAG-web}, and RAG-mix, where RAG-mix used \textsc{RAG-base} for SAQ and \textsc{RAG-web} for MCQ, all using the prompt RP-v1.





\section{Conclusions and Future Work}
In this system report, we described our FLANS system architecture design and model development for SemEval Shared Task 7 ``Everyday Knowledge Across Diverse Languages and Cultures''. 
For sustainable and secure AI development, we explored retrieval-augmented generation (RAG) using open-sourced smaller LLMs and Wikipedia extracted local knowledge base, together with online live search engine using language and model routing. 
We shared our detailed prompts developed and our codes for open-sourced research purposes.
Future work includes improving knowledge base quality control, expanding language coverage, and further optimizing small model performance.

\paragraph{Disclaimer}
The opinions and conclusions expressed in this paper are those of the authors and do not necessarily reflect the views of Insilico Medicine AI Limited.


\bibliography{custom}

@article{myung2024blend,
  title={Blend: A benchmark for llms on everyday knowledge in diverse cultures and languages},
  author={Myung, Junho and Lee, Nayeon and Zhou, Yi and Jin, Jiho and Putri, Rifki and Antypas, Dimosthenis and Borkakoty, Hsuvas and Kim, Eunsu and Perez-Almendros, Carla and Ayele, Abinew Ali and others},
  journal={Advances in Neural Information Processing Systems},
  volume={37},
  pages={78104--78146},
  year={2024}
}

@inproceedings{romero2025manchester,
  title={The Manchester Bees at PerAnsSumm 2025: Iterative Self-Prompting with Claude and o1 for Perspective-aware Healthcare Answer Summarisation},
  author={Romero, Pablo and Ren, Libo and Han, Lifeng and Nenadic, Goran},
  booktitle={Proceedings of the Second Workshop on Patient-Oriented Language Processing (CL4Health)},
  pages={340--348},
  year={2025}
}

@inproceedings{hu2023lora,
  title={LoRA: Low-Rank Adaptation of Large Language Models},
  author={Hu, Edward J. and Shen, Yelong and Wallis, Phillip and Allen-Zhu, Zeyuan and Li, Yuanzhi and Wang, Shean and Wang, Lu and Chen, Weizhu},
  booktitle={International Conference on Learning Representations (ICLR)},
  year={2023},
  url={https://openreview.net/forum?id=H1eA7AEtvS}
}

@inproceedings{jiao-etal-2020-tinybert,
    title = "{T}iny{BERT}: Distilling {BERT} for Natural Language Understanding",
    author = "Jiao, Xiaoqi  and
      Yin, Yichun  and
      Shang, Lifeng  and
      Jiang, Xin  and
      Chen, Xiao  and
      Li, Linlin  and
      Wang, Fang  and
      Liu, Qun",
    editor = "Cohn, Trevor  and
      He, Yulan  and
      Liu, Yang",
    booktitle = "Findings of the Association for Computational Linguistics: EMNLP 2020",
    month = nov,
    year = "2020",
    address = "Online",
    publisher = "Association for Computational Linguistics",
    url = "https://aclanthology.org/2020.findings-emnlp.372/",
    doi = "10.18653/v1/2020.findings-emnlp.372",
    pages = "4163--4174"
}

@inproceedings{strubell2019energy,
  title={Energy and policy considerations for deep learning in NLP},
  author={Strubell, Emma and Ganesh, Ananya and McCallum, Andrew},
  booktitle={Proceedings of the 57th annual meeting of the association for computational linguistics},
  pages={3645--3650},
  year={2019}
}

@article{kaplan2020scaling,
  title={Scaling laws for neural language models},
  author={Kaplan, Jared and McCandlish, Sam and Henighan, Tom and Brown, Tom B and Chess, Benjamin and Child, Rewon and Gray, Scott and Radford, Alec and Wu, Jeffrey and Amodei, Dario},
  journal={arXiv preprint arXiv:2001.08361},
  year={2020}
}

@article{han2024neuralMT,
  title={Neural machine translation of clinical text: an empirical investigation into multilingual pre-trained language models and transfer-learning},
  author={Han, Lifeng and Gladkoff, Serge and Erofeev, Gleb and Sorokina, Irina and Galiano, Betty and Nenadic, Goran},
  journal={Frontiers in Digital Health},
  volume={6},
  pages={1211564},
  year={2024},
  publisher={Frontiers Media SA}
}

@inproceedings{zhang-etal-2025-culturesynth,
    title = "{C}ulture{S}ynth: A Hierarchical Taxonomy-Guided and Retrieval-Augmented Framework for Cultural Question-Answer Synthesis",
    author = "Zhang, Xinyu  and
      Zhang, Pei  and
      Luo, Shuang  and
      Tang, Jialong  and
      Wan, Yu  and
      Yang, Baosong  and
      Huang, Fei",
    editor = "Christodoulopoulos, Christos  and
      Chakraborty, Tanmoy  and
      Rose, Carolyn  and
      Peng, Violet",
    booktitle = "Findings of the Association for Computational Linguistics: EMNLP 2025",
    month = nov,
    year = "2025",
    address = "Suzhou, China",
    publisher = "Association for Computational Linguistics",
    url = "https://aclanthology.org/2025.findings-emnlp.552/",
    doi = "10.18653/v1/2025.findings-emnlp.552",
    pages = "10448--10467",
    ISBN = "979-8-89176-335-7"
}

@inproceedings{faruk2025adab,
title={{ADAB}: A Culturally-Aligned Automated Response Generation Framework for Islamic App Reviews by Integrating {ABSA} and Hybrid {RAG}},
author={K.M.Tahlil Mahfuz Faruk and Mushfiqur Rahman Talha and H. M. Kawsar Ahamad and Mohammad Galib Shams and Nabil Mosharraf Hossain and Syed Rifat Raiyan and Md Kamrul Hasan and Hasan Mahmud and Riasat Islam},
booktitle={5th Muslims in ML Workshop co-located with NeurIPS 2025},
year={2025},
url={https://openreview.net/forum?id=PnWmDdwTXE}
}

@inproceedings{abdelaziz2025arabic,
  title={Arabic Mental Health Question Answering: A Multi-Task Approach with Advanced Retrieval-Augmented Generation},
  author={AbdelAziz, Abdelaziz Amr and Youssef, Mohamed Ahmed and Koritam, Mamdouh Mohamed and Eldeeb, Marwa and Hussein, Ensaf},
  booktitle={Proceedings of The Third Arabic Natural Language Processing Conference: Shared Tasks},
  pages={192--197},
  year={2025}
}

@inproceedings{joshua2024improving-african,
title={Improving Question-Answering Capabilities in Large Language Models Using Retrieval Augmented Generation ({RAG}): A Case Study on Yoruba Culture and Language},
author={Adejumobi Monjolaoluwa Joshua},
booktitle={5th Workshop on African Natural Language Processing},
year={2024},
url={https://openreview.net/forum?id=ZTt8f8ILOb}
}

@article{alan2025improving-isslamic,
  title={Improving LLM Reliability with RAG in Religious Question-Answering: MufassirQAS},
  author={Alan, Ahmet Yusuf and Karaarslan, Enis and Ayd{\i}n, {\"O}mer},
  journal={Turkish Journal of Engineering},
  volume={9},
  number={3},
  pages={544--559},
  year={2025},
  publisher={Murat YAKAR}
}

@inproceedings{li-etal-2025-multilingual,
    title = "Multilingual Retrieval Augmented Generation for Culturally-Sensitive Tasks: A Benchmark for Cross-lingual Robustness",
    author = "Li, Bryan  and
      Luo, Fiona  and
      Haider, Samar  and
      Agashe, Adwait  and
      Li, Siyu  and
      Liu, Runqi  and
      Miao, Miranda Muqing  and
      Ramakrishnan, Shriya  and
      Yuan, Yuan  and
      Callison-Burch, Chris",
    editor = "Che, Wanxiang  and
      Nabende, Joyce  and
      Shutova, Ekaterina  and
      Pilehvar, Mohammad Taher",
    booktitle = "Findings of the Association for Computational Linguistics: ACL 2025",
    month = jul,
    year = "2025",
    address = "Vienna, Austria",
    publisher = "Association for Computational Linguistics",
    url = "https://aclanthology.org/2025.findings-acl.219/",
    doi = "10.18653/v1/2025.findings-acl.219",
    pages = "4215--4241",
    ISBN = "979-8-89176-256-5"
}

@inproceedings{joseph2024retrieval-refugee,
  title={Retrieval-Augmented LLMs for Culturally Sensitive Learning Content in Refugee Education},
  author={Joseph, Tibakanya and Hellen, Nakayiza and Marvin, Ggaliwango},
  booktitle={Congress on Intelligent Systems},
  pages={425--442},
  year={2024},
  organization={Springer}
}

@article{lee2025evaluating,
  title={Evaluating cultural knowledge processing in large language models: a cognitive benchmarking framework integrating retrieval-augmented generation},
  author={Lee, Hung-Shin and Chang, Chen-Chi and Chen, Ching-Yuan and Hsu, Yun-Hsiang},
  journal={The Electronic Library},
  pages={1--22},
  year={2025},
  publisher={Emerald Publishing Limited}
}

@article{ren2025paisp,
  title={Malei at MultiClinSUM: Summarisation of Clinical Documents using Perspective-Aware Iterative Self-Prompting with LLMs},
  author={Ren, Lei and Ng, Y. M. and Han, L.},
  journal={arXiv preprint arXiv:2509.07622},
  year={2025}
}

\clearpage

\appendix

\section{\textsc{RAG-base} and \textsc{RAG-web}}
\begin{figure*}[htbp]
    \centering
    \includegraphics[width=0.8\textwidth]{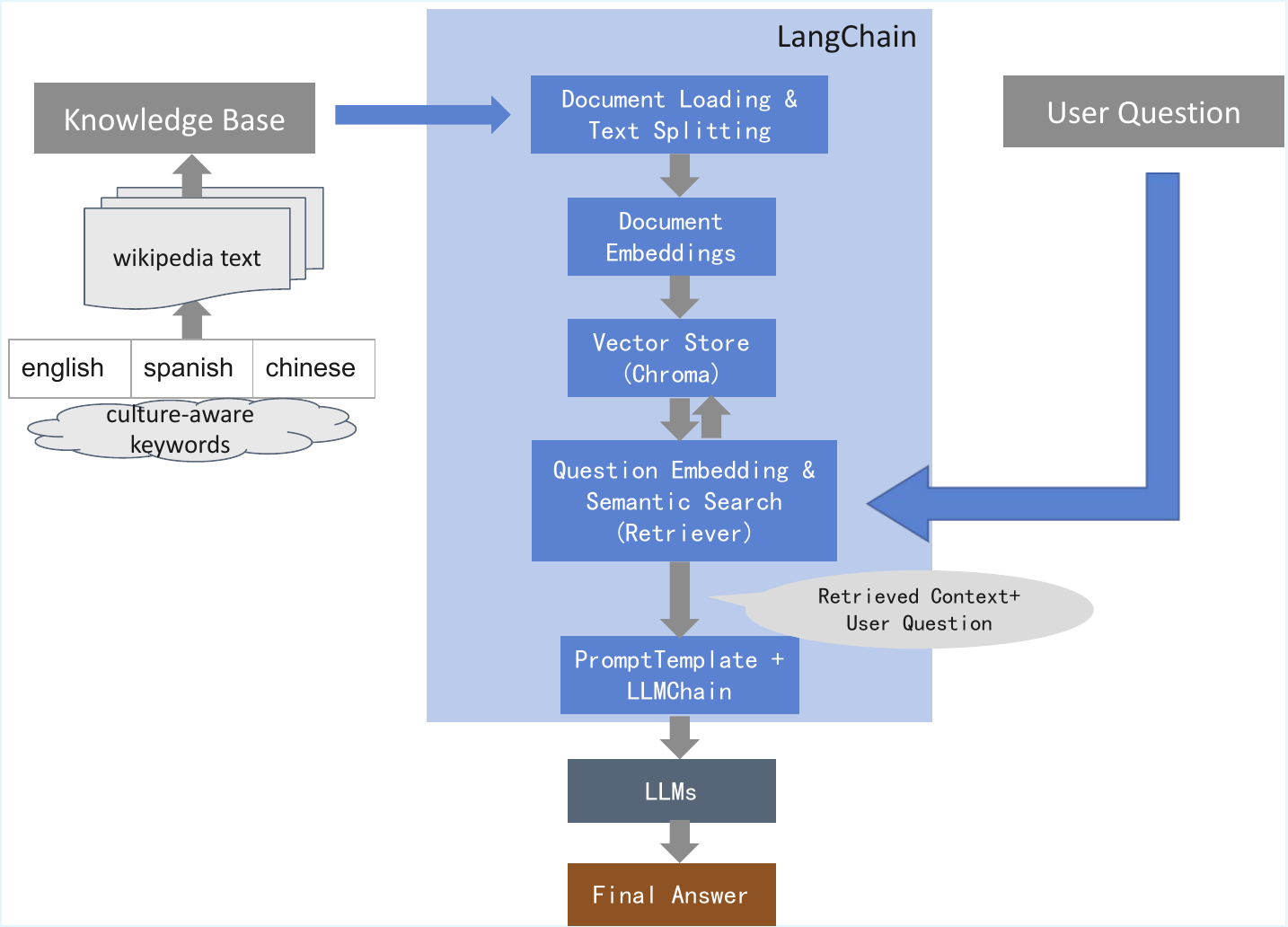}
    \caption{\textsc{RAG-base} pipeline using smaller LLMs favoring Gemma3.4b - keywords based KE - then land to KB}
    \label{fig:rag_pipeline_shiran}
\end{figure*}

\begin{figure*}[htbp]
    \centering
    \includegraphics[width=\textwidth]{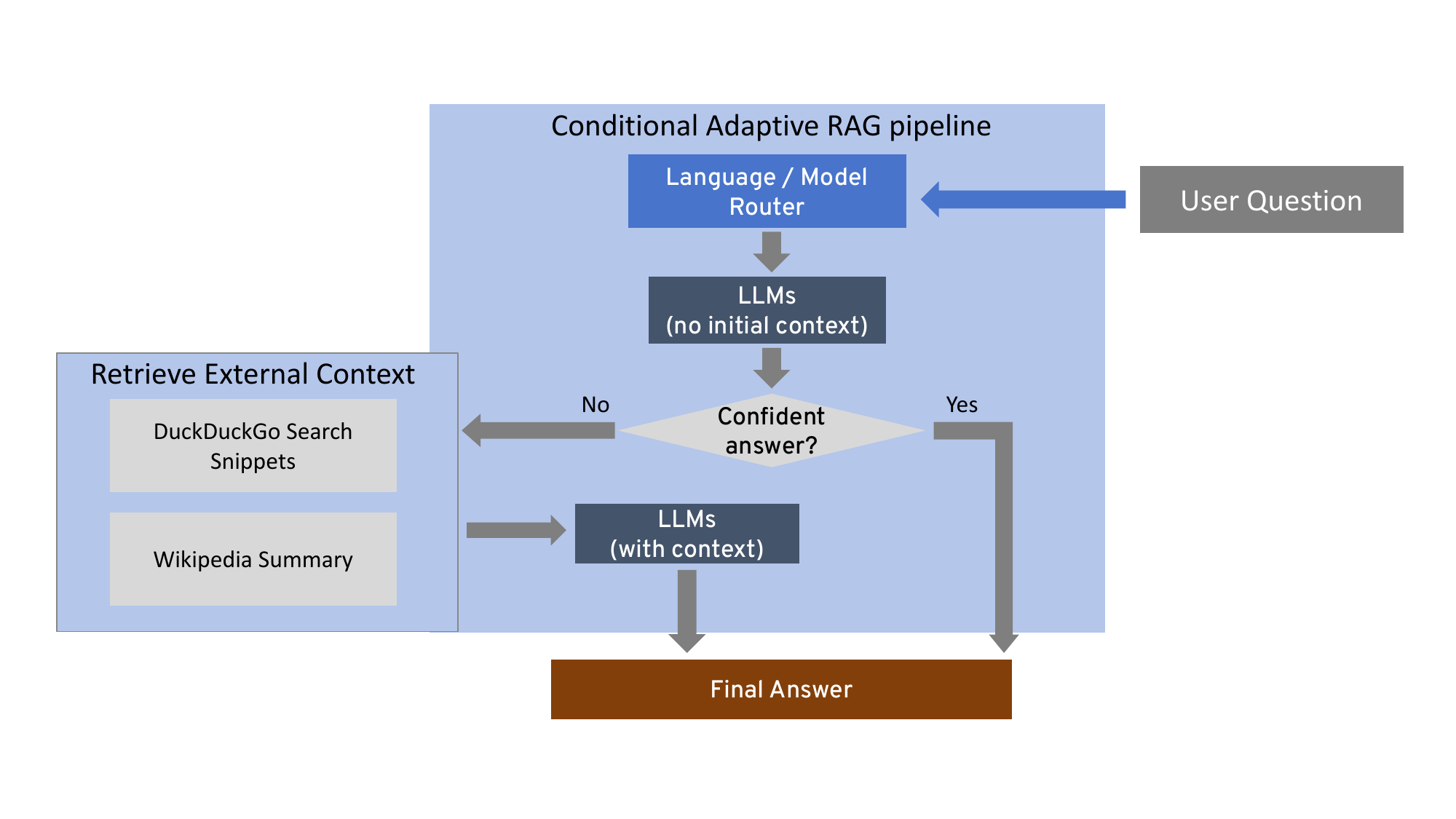}
    \caption{\textsc{RAG-web} pipeline using mistral:7b and deepseek-llm:67b (bigger) - Conditional adaptive RAG}
    \label{fig:rag_pipeline_lillia}
\end{figure*}

The two RAG systems we developed before merging are presented in Figure \ref{fig:rag_pipeline_shiran} and \ref{fig:rag_pipeline_lillia}.

The steps to run \textsc{RAG-base}:
\begin{itemize}
    \item 1st: run prompt; there is no output but just to set it up.
    \item 2nd: run RAG; it might lead to some output.
    \item 3rd: print answer from LLM-w-RAG.
\end{itemize}

\subsection{\textsc{RAG-web} Design Motivation}
The BLEnD shared task tests culturally grounded and country-specific knowledge in a wide variety of languages and geographic locations. A large proportion of the questions involve common practices, local customs, institutions, and/or facts that are particular to a geographic area and are not expected to be encoded in the parametric memory of a small language model. On the other hand, the BLEnD shared task does not allow task fine-tuning, which calls for an inference strategy that boosts grounding while maintaining low computational costs and reproducibility.

In light of the above-mentioned issues with the BLEnD shared task, we have devised a cascaded retrieval-augmented inference strategy that is optimized for local execution. Instead of depending on a single source of information, the model combines different channels of evidence with a fallback strategy: \textbf{web search} with low overhead, local knowledge base with country-specific information from Wikipedia (optional), and responses from a parametric model when applicable. The model prioritizes precision and format correctness over recall and explicitly allows abstention through a <NO\_ANSWER> output when evidence cannot be established. The final model strikes a balance between local search and quality of answers due to a trade-off observed during development: local search improves recall but hurts the quality of answers for culturally nuanced short answers.

    \subsection{\textsc{RAG-web} KB}

For each country included in BLEnD, a brief set of highly informative Wikipedia pages regarding country overviews, culture, cuisine, tourism, history, etc., was automatically compiled. This collection was then split into overlapping text segments using a character-based text splitter, which ensures semantic coherence. Each text segment was then embedded by \texttt{OllamaEmbeddings} (Mistral-based) before being stored in a country-specific Chroma vector database, \texttt{db\_\{country\}}.

This structure enables efficient semantic retrieval, ensuring that culturally relevant data remains geographically relevant. Notably, the local knowledge bases are used as optional evidence sources, not as a source of ground truth, with access restricted at runtime.
\subsection{\textsc{RAG-web} vs Traditional RAG}
While the methodology takes its cue from conventional RAG frameworks, the proposed system differs in that it does not strictly follow a one-pass retrieval and generation strategy, as in the case of conventional RAGs, and instead adopts a cascaded inference strategy that is more suitable for the BLEnD task. Each country in the dataset is associated with an optional local knowledge base, implemented as a separate Chroma vector database. The databases are built on a curated subset of relevant Wikipedia pages, including those related to national overview, culture, cuisine, tourism, and history.

In terms of its inference strategy, the proposed system follows a multi-stage decision process. First, the question ID is tokenized and analyzed to determine the language and country code, which in turn determines the language model and, if enabled, the country-specific local knowledge base. The system next determines if a direct model-only response is appropriate, a strategy primarily reserved for encyclopedic types of queries and not culturally informed short-answer queries, in which case an unguided generation process is not reliable.

In terms of retrieval, if evidence is requisite, the system uses a prioritized cascaded strategy. First, web search using \texttt{DuckDuckGo} is attempted, as this search engine is more likely to return concrete and contextually relevant information snippets relevant to everyday cultural queries (we consider the top-8 candidates). 
If this fails, the system will attempt a semantic similarity search using the country-specific webpage from Wikipedia summary (the free text in the top the of the page).

At each stage, retrieved evidence is injected into the prompt and the model is asked to answer only if the answer is explicitly supported by the context. If no stage produces a valid answer, the system outputs \texttt{<NO\_ANSWER>}.

\subsection{\textsc{RAG-web} inference}
Our final stage of inference utilizes a combination of retrieval from multiple external sources with dynamic routing of the language model. Depending on the type of query and availability, the system may retrieve supporting evidence from live web search, country-specific local knowledge bases, or Wikipedia summaries.

In the presence of local knowledge bases, queries are embedded in the same vector space as country-specific \texttt{Chroma} databases to retrieve semantically relevant cultural documents. Local knowledge retrieval, however, is not a necessary component, as empirical analysis showed that local retrieval can add noise to culturally grounded short-answer questions.

For most queries, web search through \texttt{DuckDuckGo} is preferred, especially when queries are about general practices, cuisine, or institutions. Wikipedia summaries are used as a fallback option when queries are encyclopedic or historical in nature. Retrieved passages are embedded into a prompt template, with the model being asked to respond only when the answer is explicitly supported by the evidence provided.

\section{Detailed Prompts}
\label{sec:appendix}

This appendix presents the three prompt variants used in our prompt ablation study.
\subsection{Minimal Prompt (MP)}
\paragraph{}
\begin{Verbatim}[breaklines=true, breakanywhere=true, fontsize=\small]
"""
Our goal is to give one short, correct answer for each question in its original language.

"""
\end{Verbatim}

\subsection{Refined Prompt-v1 (RP-v1): Instructions + Persona}
\paragraph{}
\begin{Verbatim}[breaklines=true, breakanywhere=true, fontsize=\small]
"""
You are a factual multilingual assistant for a question-answering benchmark.
Your goal is to give one short, correct answer for each question in its original language.

Instructions:
- Read the question carefully.
- Respond ONLY with the concise answer — a word, number, name, or short phrase.
- Do not include explanations, reasoning, labels, or extra words.
- If the question asks for a person, place, or date, give only that entity.
- Keep the answer in the SAME language as the question
  (Chinese → Chinese, English → English, Spanish → Spanish).

Now answer the following question.

"""
\end{Verbatim}

\subsection{Refined Prompt-v2 (RP-v2): Persona + Perspective-aware + CoTs}
\paragraph{}
\begin{Verbatim}[breaklines=true, breakanywhere=true, fontsize=\small]
"""
You are a factual, multilingual assistant for a question-answering benchmark.
Your goal is to produce ONE short, correct answer for each question in its original language.
Before answering, please think about the task:

1. Question Analysis:
- Identify the language of the question.
- Identify what type of information is being asked (person, place, date, object, concept, number, or other).

2. Information Focus:
- Determine the single factual element required to answer the question.
- Ignore any irrelevant or descriptive details.

3. Answer Strategy:
- Recall general world knowledge relevant to the question.
- Prefer the most standard, widely accepted answer.
- Avoid over-specific or explanatory phrasing.

4. Error Avoidance:
- Do NOT include explanations, reasoning, or extra words.
- Do NOT translate or restate the question.
- Do NOT include multiple candidates or alternatives.

After this internal analysis, provide ONLY the final answer.

Answering Rules:
- Output a single short answer (a word, name, number, or short phrase).
- Keep the answer in the SAME language as the question.
- Do not include labels, punctuation, or additional text.
- If uncertain, give your best plausible short answer based on general knowledge.

Now answer the following question.

"""
\end{Verbatim}

\end{document}